\documentclass[runningheads]{waica}
\usepackage[T1]{fontenc}
\usepackage{graphicx}
\usepackage{caption}   
\usepackage{subcaption} 
\usepackage{booktabs}  
\usepackage{graphicx}  

\PassOptionsToPackage{table}{xcolor}
\usepackage[utf8]{inputenc} 
\usepackage[T1]{fontenc}    
\usepackage{hyperref}       
\usepackage{url}            
\usepackage{booktabs}       
\usepackage{amsfonts}       
\usepackage{nicefrac}       
\usepackage{microtype}      
\usepackage{wrapfig}      
\usepackage{lipsum}       
\usepackage{booktabs}     
\usepackage{caption}      
\usepackage{graphicx}
\usepackage{enumitem}
\usepackage{natbib}  
\usepackage{spverbatim}
\usepackage[many]{tcolorbox}
\usepackage{lipsum} 
\usepackage{fancybox}
\usepackage{verbatim}
\usepackage{fancyvrb}
\usepackage{xspace}
\usepackage{svg} 
\definecolor{mygreen}{RGB}{50,100,50} 
\definecolor{bggreen}{RGB}{220,240,220} 
\definecolor{deepblue}{RGB}{0, 51, 102}   
\definecolor{lightblue}{RGB}{173, 216, 230}  
\definecolor{oceanblue}{RGB}{0, 102, 204}   
\definecolor{skyblue}{RGB}{135, 206, 250}   
\definecolor{iceblue}{RGB}{173, 216, 230}   
\definecolor{darkice}{RGB}{65, 105, 225}    
\definecolor{techblue}{RGB}{51, 153, 255}   
\definecolor{whitebg}{RGB}{245, 255, 255}   

\definecolor{lightgray}{RGB}{245,245,245} 

\usepackage{listings}

\definecolor{codegreen}{rgb}{0,0.6,0}
\definecolor{codegray}{rgb}{0.5,0.5,0.5}
\definecolor{codepurple}{rgb}{0.58,0,0.82}
\definecolor{backcolor}{rgb}{0.95,0.95,0.95}

\lstdefinestyle{mystyle}{
    backgroundcolor=\color{backcolor},
    commentstyle=\color{codegreen},
    keywordstyle=\color{magenta},
    stringstyle=\color{codepurple},
    basicstyle=\footnotesize\ttfamily,
    breakatwhitespace=false,
    breaklines=true,
    captionpos=b,
    keepspaces=true,
    numbers=left,
    numbersep=5pt,
    numberstyle=\tiny\color{codegray},
    showspaces=false,
    showstringspaces=false,
    showtabs=false,
    tabsize=2,
    frame=single
}

\usepackage{listings}

\definecolor{lightgray}{RGB}{245,245,245}
\definecolor{codegreen}{rgb}{0,0.6,0}
\definecolor{codegray}{rgb}{0.5,0.5,0.5}
\definecolor{codepurple}{rgb}{0.58,0,0.82}
\definecolor{backcolour}{RGB}{249,249,249}

\PassOptionsToPackage{square}{natbib}
\setcitestyle{numbers,square}

\newcommand{\sandbox}{\textsc{InternBootcamp}}%
\newcommand{\eval}{\textsc{Bootcamp-Eval}}%

\newcommand{\llmname}[1]{{\fontfamily{pcr}\selectfont {#1}}\xspace}
\begin{document}
\title{InternBootcamp:
Boosting LLM Reasoning with Verifiable Task Scaling}
%


%
%
%

\makeatletter
\newcommand{\equalmark}{\textsuperscript{\@fnsymbol{1}}}
\newcommand{\corrmark}{\textsuperscript{\@fnsymbol{2}}}
\makeatother

\author{
Peiji Li\inst{1,2}\thanks{Equal contribution.} \and
Jiasheng Ye\inst{1,2}\equalmark \and
Yongkang Chen\inst{1}\equalmark \and
Linyang Li\inst{1}\thanks{Corresponding author.} \and
Yichuan Ma\inst{1,2} \and
Zijie Yu\inst{1} \and
Ganqu Cui\inst{1} \and
Haozhan Li\inst{1} \and
Jiacheng Chen\inst{1} \and
Chengqi Lyu\inst{1} \and
Wenwei Zhang\inst{1} \and
Qipeng Guo\inst{1} \and
Dahua Lin\inst{1} \and
Bowen Zhou\inst{1}\corrmark \and
Kai Chen\inst{1}\corrmark
}

\authorrunning{P. Li et al.}

\institute{
Shanghai AI Laboratory \\
\email{\{lipeiji, lilinyang, chenkai\}@pjlab.org.cn} \and
Fudan University \\
\email{24110240171@m.fudan.edu.cn} \\[0.5em]
}

\maketitle              
\begin{abstract}
Large language models (LLMs) have revolutionized artificial intelligence by enabling complex reasoning capabilities. While recent advancements in reinforcement learning (RL) have primarily focused on domain-specific reasoning tasks (e.g., mathematics or code generation), real-world reasoning scenarios often require models to handle diverse and complex environments that narrow-domain benchmarks cannot fully capture.
To address this gap, we present \sandbox, an open-source framework comprising 1000+ domain-diverse task environments specifically designed for LLM reasoning research. 
Our codebase offers two key functionalities: (1) automated generation of unlimited training/testing cases with configurable difficulty levels, and (2) integrated verification modules for objective response evaluation. 
These features make \sandbox{} fundamental infrastructure for RL-based model optimization, synthetic data generation and model evaluation.
Although manually developing such a framework with enormous task coverage is extremely cumbersome, we accelerate the development procedure through an automated agent workflow supplemented by manual validation protocols, which enables the task scope to expand rapidly. 
With these bootcamps, we further establish \eval{}, an automatically generated benchmark for comprehensive performance assessment.
Evaluation reveals that frontier models still underperform in many reasoning tasks, while training with \sandbox{} provides an effective way to significantly improve performance, leading to our 32B model that achieves state-of-the-art results on \eval{} and excels on other established benchmarks.
In particular, we validate that consistent performance gain come from including more training tasks, namely \textbf{task scaling}, over two orders of magnitude, offering a promising route towards capable reasoning generalist.
 All data and code are publicly available.

\end{abstract}
\section{Introduction}

Large language models (LLMs) with enhanced reasoning capabilities~\cite{r1,o1,qwen3,minimaxm1, seedthinking} have demonstrated remarkable performance in complex reasoning tasks, particularly showing strong post-training adaptation potential through reinforcement learning (RL) paradigms~\cite{dapo,vapo,luffy,ttrl}. However, these advancements remain concentrated within domains like mathematics and code generation, where abundant question-answer datasets facilitate model learning. In contrast, LLMs exhibit significant performance gaps compared to human experts when handling more diverse reasoning tasks that require cross-domain knowledge integration.

Recent efforts~\cite{Chen2025EnigmataSL,synlogic,Shi2025KORGymAD} have focused on synthesizing verifiable logical reasoning and puzzle datasets compatible with the Reinforcement Learning with Verifiable Rewards (RLVR). While promising, these works typically contain only dozens of narrowly scoped tasks, failing to comprehensively assess cross-domain reasoning capabilities required for real-world applications.
\begin{figure}[t] 
        \centering
        \includegraphics[width=\linewidth]{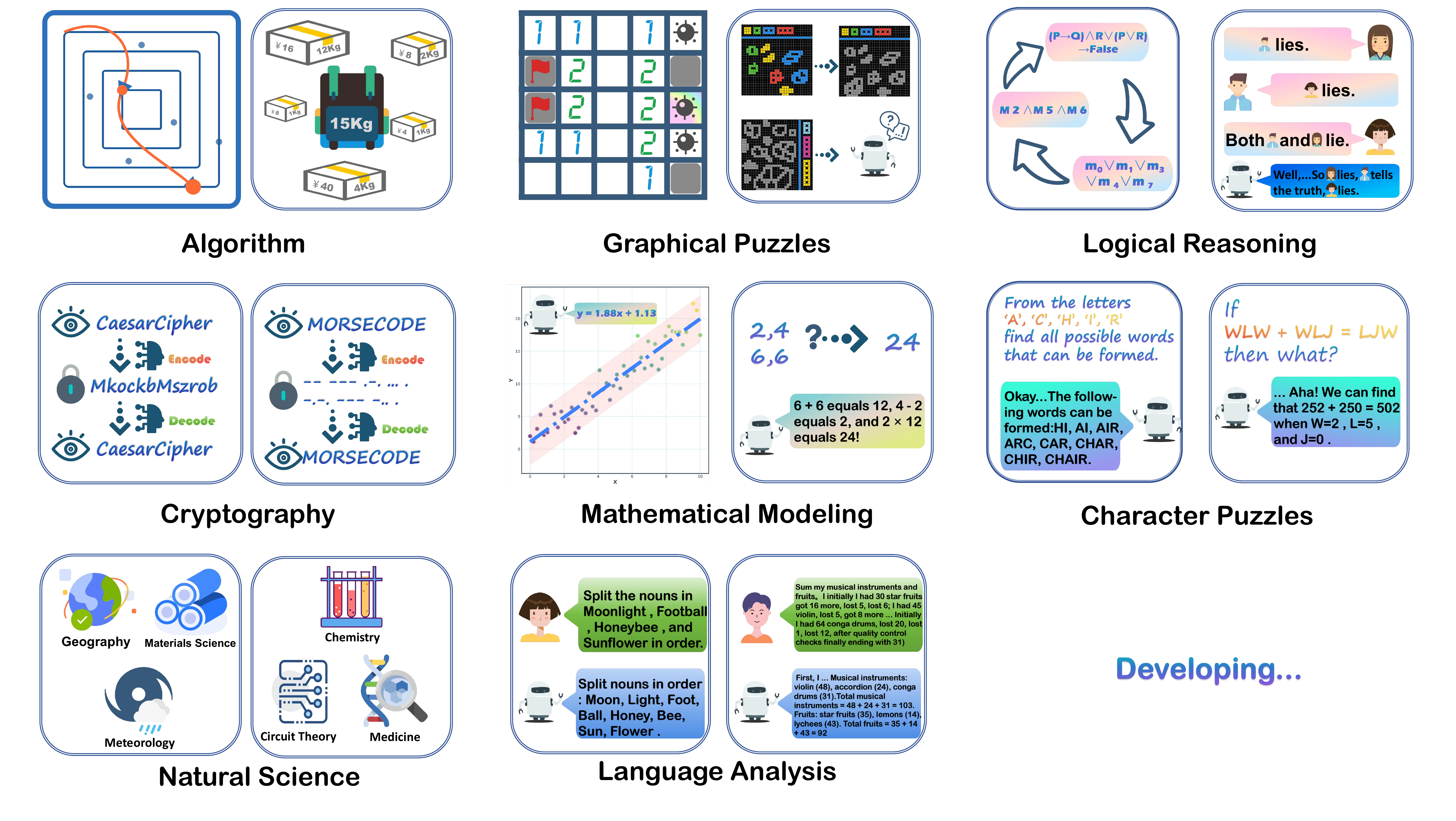}
        \caption{Overview of reasoning tasks supported by \sandbox{}, covering over 1000 tasks across 8 categories from real-world reasoning scenarios.}
        \label{fig:bootcamp_example}
\end{figure}

To address these limitations, we present \sandbox, a large-scale interactive training framework enabling LLMs to learn verified solutions across diverse general reasoning tasks. Unlike existing works that predominantly focus on mathematical/code-centric domains or constrained logical puzzles, our framework spans 1000+ reasoning tasks categorized into 8 broad domains, including Algorithm, Language Analysis and Natural Science, which is described in Figure~\ref{fig:bootcamp_example}. By prioritizing real-world relevance through systematically constructed environments, \sandbox{} advances multi-domain reasoning capabilities beyond narrow-specialized benchmarks.

As a fundamental feature, \sandbox{} provides a general framework for diverse reasoning tasks, enabling seamless integration with other frameworks for model evaluation, Reinforcement Learning ~\cite{xtuner,verl} or synthetic data generation. 
In \sandbox{}, each task is encapsulated as a Bootcamp class with configurable parameters to control difficulty levels. 
These classes implement standardized interfaces for two core functionalities: (1) infinite question generation through parameterized templates, and (2) automated verification of LLM-generated responses via built-in verifiers. 
These unified interfaces enable users of \sandbox{} to manage numerous tasks simultaneously with little engineering burden as the number of tasks increases.

Building upon our framework, we begin by manually curate 118 reasoning tasks across domains, on which we establish \eval{} - a benchmark dataset containing 9,232 samples across the 118 reasoning tasks.
\eval{} enables a comprehensive assessment of cross-domain reasoning capabilities in LLMs.
Through evaluation on it, we confirm that merely training models on a narrow scope of tasks such as math indeed hardly facilitates the improvement of reasoning performance on more diverse domains, thus necessitating expanding task scopes in developing reasoning models.

\begin{figure}[!b]
  \vspace{-20pt}
    \centering
    \includegraphics[width=\linewidth]{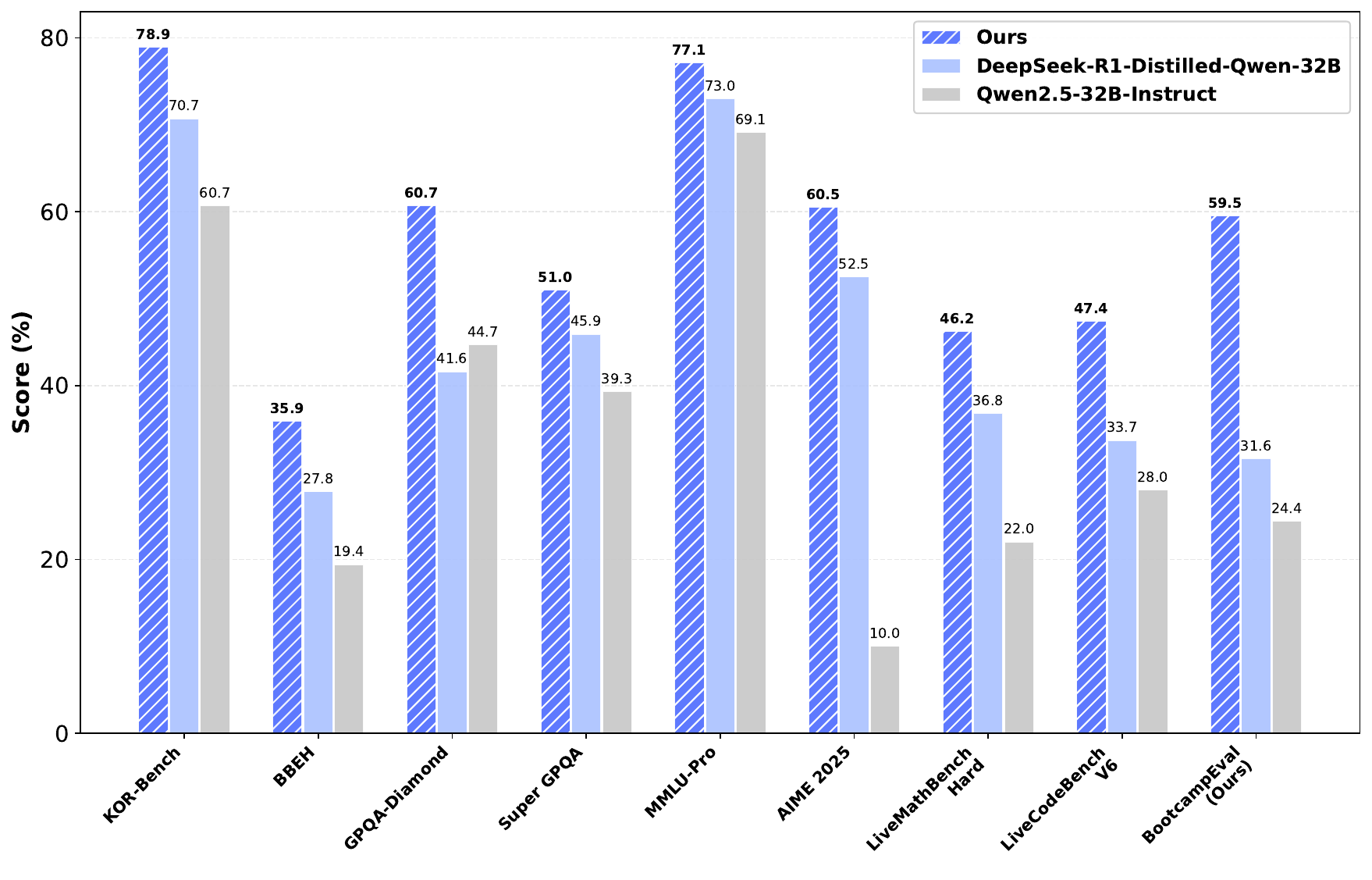}
    \caption{Performance comparison across multiple reasoning benchmarks. Our model, trained with reinforcement learning on \sandbox{} tasks from Qwen2.5-32B-Instruct, achieves superior reasoning performance and generalization compared to others.}

    \label{fig:overall}
    \vspace{-20pt}
\end{figure}

As a result, we try to incorporate more tasks and investigate into the relationship between LLM reasoning capabilities and training task scaling. To achieve this, we developed an \textbf{automatic agent workflow} for Bootcamp class synthesis based on our unified interface. This pipeline involves: (1) collecting task specifications, (2) leveraging strong reasoning models (e.g., \llmname{Deepseek-R1}~\cite{r1}) to generate Bootcamp classes in a evolutionary manner, (3) filtering non-executable or anomalous code through self-consistent unittests, and (4) human review. This systematic approach eventually expanded our task collection to over 1000 Bootcamp classes. The sufficient diversity and scale of these tasks enabled our \textbf{task scaling} experiments, which  demonstrate that increasing reasoning task quantity during training not only enhances reasoning performance but also improves training efficiency. By providing a rich and scalable environment for models to engage with diverse reasoning challenges, our framework supports the potential of \textit{experiential learning}~\cite{eraofexperience}, where agents improve through feedback from the environment—a key pathway toward more general and capable intelligence. Figure~\ref{fig:overall} demonstrates that our 32B model, trained on the full task set from \sandbox{}, exhibits strong generalization across multiple reasoning benchmarks.

We summarize the key insights from experimental observations:

\begin{itemize}[leftmargin=15pt, labelwidth=*, labelsep=5pt, parsep=0pt, topsep=0pt]
\item \textbf{Scalable task synthesis enables broad experiential learning:} Our automated agent workflow demonstrates that large-scale, diverse reasoning environments can be effectively synthesized via iterative, evolutionary methods. This opens the door to training agents on a continuous stream of novel tasks—aligning with the principles of experiential learning.

\item \textbf{Generalization emerges from cross-task exposure:} LLMs develop stronger reasoning generalization and emergent abilities not through deep specialization in narrow domains, but by learning across a wide spectrum of reasoning tasks. This supports the view that intelligence arises from breadth of experience, not just depth of imitation.

\item \textbf{Task scaling improves both performance and efficiency:} In our RLVR experiments, increasing the number of training tasks significantly boosts both final performance and learning efficiency—a phenomenon we term \textbf{task scaling}. This suggests that exposing models to more diverse experiences accelerates and strengthens their reasoning development.
\end{itemize}

Finally, our key contributions are as follows.

\begin{itemize}[leftmargin=15pt, labelwidth=*, labelsep=5pt, parsep=0pt, topsep=0pt]
    \item We propose \sandbox{}, the first large scale extensible library of environments for training large reasoning models, supporting over 1000 general reasoning tasks across 8 domains
    \item We construct \eval{}, a comprehensive benchmark containing 9,232 samples across 118 cross-domain reasoning tasks to extensively evaluate the generality of reasoning models.
    \item We empirically demonstrate the effectiveness of \textbf{Task Scaling} in reinforcement learning, showing that scaling up training tasks improves both reasoning performance and efficiency, providing a practical pathway toward reasoning generalists.
\end{itemize}

\section{Related Works}

\subsection{Reasoning Studies in LLMs}

LLMs demonstrate remarkable ability in reasoning benchmarks (e.g., MMLU~\cite{hendrycks2020measuring}, MATH~\cite{MATH}, AIME~\cite{aime}, BBEH~\cite{BBEH}, KOR~\cite{korbench}), covering both knowledge-related and knowledge-orthogonal tasks.
Such ability is developed based on a series of research initiated from chain-of-thought reasoning~\cite{COT}, allowing autoregressive language models to explore step by step and solve reasoning tasks in a manner similar to human thoughts.
Such technology is quickly studied in later research, exemplified by self-consistency~\cite{selfconsistency} and massive prompt-engineering approaches~\cite{xu2023wizardlm,yu2023metamath}
After these CoT-enhancing studies, the generalization ability of chain-of-thoughts is widely studied in various aspects, such as test-time-scaling with long CoT systems through Reinforcement Learning with Verifiable Rewards (RLVR) like \llmname{OpenAI-O1}~\cite{o1} and \llmname{Deepseek-R1}~\cite{r1},
tree-search-based path-exploring methods such as Tree-of-Thoughts~\cite{yao2024tree} or Monte-Carlo Tree Search based methods~\cite{alphamath,li2025fastmcts}.
These works aim to explore how to increase the generalization ability in LLM reasoning tasks.

\subsection{Environments for General Reasoning Tasks}

Environments have proven effective testbeds for advancing deep learning through reinforcement learning frameworks~\cite{AlphaGO,alphagozero,alphazero}, exemplified by breakthroughs in Go and chess. Recent studies extend this paradigm to LLM reasoning enhancement via interactive environments in specialized domains~\cite{alphageometry, alphafold, AlphaEvolve, wang2023voyager}. However, these works predominantly target single-task scenarios, which constrain their cross-domain generalization potential. Alternative approaches focus on logical reasoning through puzzle-solving or gameplay environments~\cite{Chen2025EnigmataSL, synlogic, Shi2025KORGymAD, gamebot,tong2025code2logic}, offering multi-task data generation and verification paradigms distinct from math/code-centric benchmarks. Despite these advances, existing frameworks still exhibit limited domain coverage and struggle to generalize to real-world reasoning tasks. To address these limitations, we present \sandbox{} – a systematic framework that automatically constructs hundreds of diverse reasoning tasks spanning multiple domains through an automated agent pipeline.

\section{InternBootcamp: A Large-Scale Task Library for Training LRMs}
\label{sec:InternBootcamp}

In this section, we present \sandbox, a library that serves as an extensible framework that provides vast environments to train and evaluate large reasoning models. 
By covering over 1000 reasoning tasks with unified interfaces to generate difficulty-controllable task samples and provide rule-based verification, we envision \sandbox{} as an infrastructure for systematically studying the effects of task mixing and task scaling in the development of reasoning models.
We detailed our considerations and techniques to enable efficient development as follows.

\begin{figure*}[htb]
    \centering
    \includegraphics[width=\linewidth]{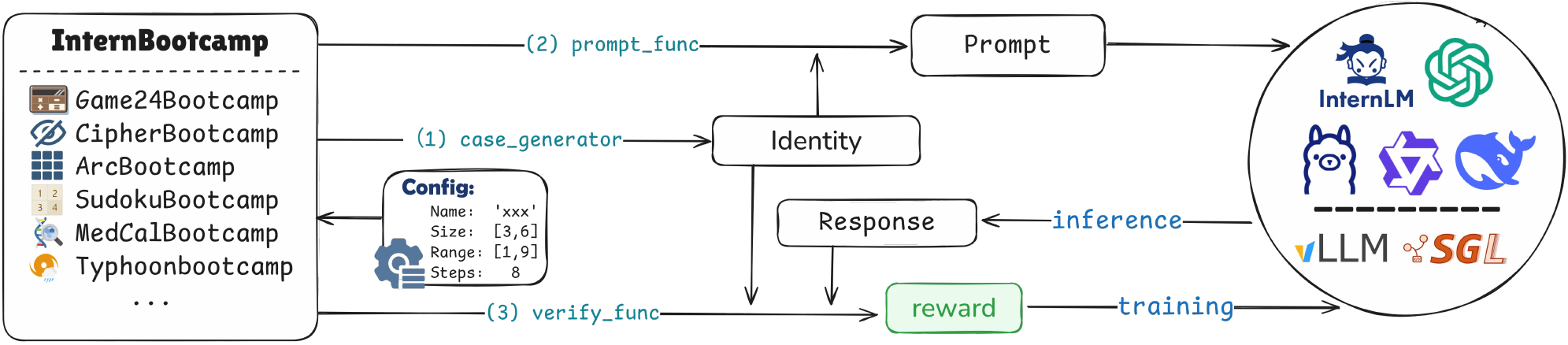}
    \caption{An overview of the framework of \sandbox{}}
    \label{fig:interface}
\end{figure*}

\subsection{Task Sources}
\label{sec:task_sources}
We draw our task collection from a diverse set of real-world and synthetic reasoning domains, designed to cover a broad spectrum of reasoning behaviors—from deductive logic in puzzles to algorithmic thinking and scientific problem-solving. Our data sources include public puzzle repositories, established reasoning benchmarks, competitive programming datasets, and scientific reasoning tasks, all carefully curated to ensure verifiability, diversity, and scalability. These tasks are systematically organized into bootcamp classes to support our experiments. A detailed breakdown of task sources and the curation process is provided in Appendix~\ref{apd:task_sources}.

\subsection{Design of the Framework}

\label{sec:framework}
Given the wide range of reasoning tasks introduced in Section~\ref{sec:task_sources}, working with them simultaneously could be cumbersome.
To address this, aligned with the needs of RLVR and data synthesis, we design \sandbox{} to provide unified interfaces for generating problem instances and offering automatic verification for these tasks.
Specifically, \sandbox{} establishes a modular architecture through the \texttt{BaseBootcamp} abstract base class.
We implement each task as a concrete Bootcamp class inheriting from \texttt{BaseBootcamp} and implements three core methods: \texttt{case\_generator} to synthesize problem instances,  \texttt{prompt\_function} to transform problem instances into text, and \texttt{verify\_function} to provide verification for a solution response to the problem instance. Additionally, each bootcamp class accepts an user-defined configuration file during initialization, which encodes hyperparameters for difficulty control and initialization. The demonstration of the framework of \sandbox{} is described in Figure \ref{fig:interface}.
For a detailed description of the framework and its implementation, see Appendix~\ref{apd:framework}.

\subsection{Automatic Agent Workflow for Large-Scale Bootcamp Synthesis}
\label{sec:autogen}

\begin{figure*}[tb]
    \centering
    \includegraphics[width=\linewidth]{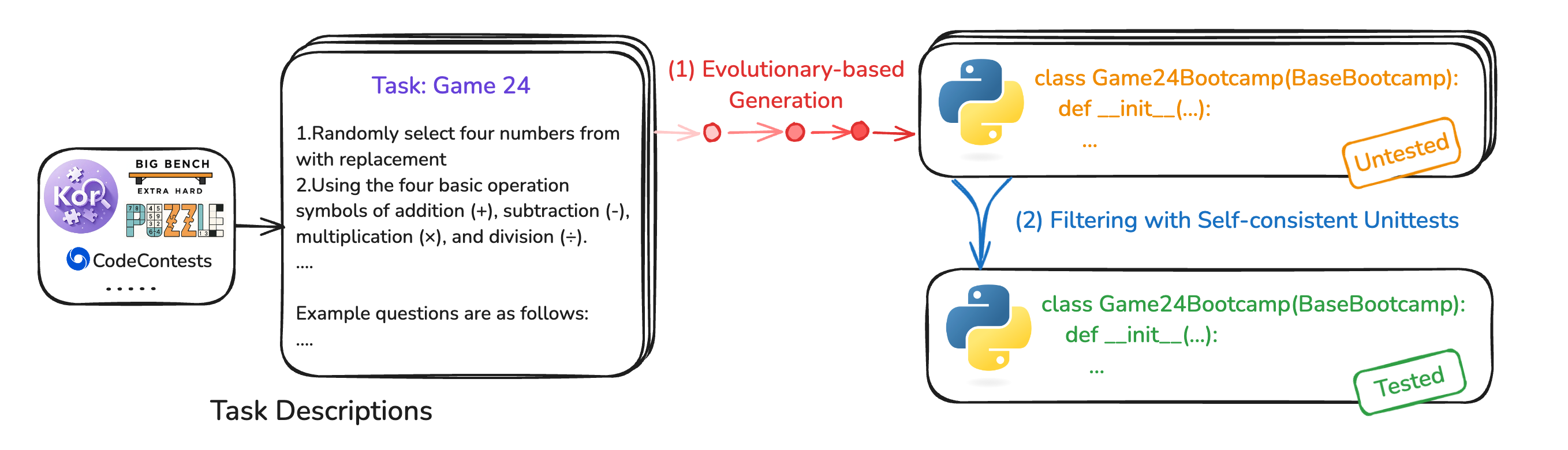}
    \caption{Illustration on the automatic agent workflow for large-scale bootcamp synthesis. We first collect diveser task descriptions and generate candidate bootcamps with an evolutionary-based method, which iteratively refine the generated bootcamp to improve the generation quality. 
    After generation, we use self-consistent unittests to evaluate a language model and filter out bootcamps with extremely high or low scores, which are probably problematic. 
    Please check Sec.~\ref{sec:autogen} for detailed descriptions on the generation and unittest filtering. }
    \label{fig:autogen}
\end{figure*}

Despite the existence of fruitful tasks, converting these tasks into Bootcamps requires tedious coding efforts.
Merely relying on human endeavors to implement all the Bootcamps is hardly possible.
Thanks to the promising progress in coding capabilities of frontier language models, we find it possible to build an agent workflow to generate the bootcamps automatically, which significantly accelerates the expansion of tasks supported by \sandbox{}.

The overview of our agent workflow is demonstrated in Figure~\ref{fig:autogen}. 
In particular, we highlight two key ingredients to ensure the quality of the automatically generated bootcamps.
\begin{itemize}[leftmargin=15pt, labelwidth=*, labelsep=5pt, parsep=0pt, topsep=0pt]
    \item First, during bootcamp generation, we adopt an evolutionary-based procedure which iteratively refine the generated bootcamps with feedback.
    \item Second, after generating the bootcamps, we use them to score a language model and filter out problematic bootcamps according to the test accuracy, namely the self-consistent unittest.
\end{itemize}
We illustrate the details of these two techniques and their effects as follows.

\paragraph{Evolutionary-based Bootcamp Generation.}

To generate a bootcamp, we provide the task description (Section~\ref{sec:task_sources}) and the predefined interfaces (Section~\ref{sec:framework}) to \llmname{Deepseek-R1} and require the model to implement the corresponding bootcamp by filling in the interfaces.
In our preliminary attempts, we find that the model tends to generate bootcamps that simplify the tasks to a large extent.
We attribute this to the characteristics of the reasoning model as we find 97.93\% (221 out of 228) of the models' thinking processes contain key words relating to the attempts of simplification\footnote{Representative key words include \textit{"to simplify"}, \textit{"let's ... temporarily"}, \textit{"might be very difficult"}.}.
Typically, many produced bootcamps can only generate a few special problem instances.

\begin{table}[htbp]
  \centering
  \begin{minipage}{0.48\textwidth}
    \centering
    \caption{ Frequency of simplification keywords in LLM thinking during bootcamp generation (228 experiments). 
      Repeated keywords in one iteration are only counted once.}
    \label{tab:simplification frequency}
    \vspace{5pt}
    \resizebox{\linewidth}{!}{ 
      \begin{tabular}{ccc}
        \toprule
        \textbf{Iteration} & \textbf{Occurrence Count} & \textbf{Frequency} \\
        \midrule
        1 &  221  & 97.93\% \\
        2 &  124  &  54.39\%  \\
        3 &  74  & 32.46\% \\
        \bottomrule
      \end{tabular}
    }
  \end{minipage}
  \hfill 
  \begin{minipage}{0.48\textwidth}
    \centering
    \caption{
      Percentage of automatically produced bootcamps that can only generate several problematic  instances. 
      We experiment with 228 attempts, each of which takes three iterations.
    }
    \label{tab:problematic frequency}
    \vspace{10pt}
    \resizebox{\linewidth}{!}{ 
      \begin{tabular}{cc}
        \toprule
        \textbf{Iteration} & \textbf{Proportion of Problematic Bootcamps} \\
        \midrule
        1 &  33 / 228\\
        2 &  19 / 228 \\
        3 &  14 / 228\\
        \bottomrule
      \end{tabular}
    }
  \end{minipage}
  \vspace{-10pt}
\end{table}

To mitigate this, we employ an iterative refinement process in which LLMs improve the bootcamps using feedback from execution\footnote{The feedback includes whether the bootcamp can be successfully executed and whether it generates only a limited range of problem instances.}. This enables the models to progressively enhance the bootcamps by incorporating details missed in earlier rounds. Statistical analysis shows that with more iterations, language models are less likely to oversimplify the generated bootcamps (Table~\ref{tab:simplification frequency}), and fewer bootcamps produce only a narrow set of problem instances(Table~\ref{tab:problematic frequency}).

\paragraph{Self-Consistent Unittest for Bootcamp Filtering.}

Although evolutionary-based generation processes enhance the quality of generated content, they still cannot guarantee the correctness of the generated bootcamps.
Therefore, it is necessary to filter out potentially problematic candidates.
In addition to checking for execution errors, our approach leverages the generated bootcamp itself to evaluate its correctness through self-consistent unittests.
Specifically, we use the bootcamp to generate problem instances and employ large language models to solve these instances.
We then verify the language models' responses using  the \texttt{verify\_function} of the bootcamp itself.
The underlying intuition is that if a bootcamp is overly simplified, the solution accuracy will be extremely high; conversely, if its implementation contains semantic errors, the accuracy will be extremely low (nearly zero).
In practice, we filter out bootcamps which produce an accuracy over 0.85 or below 0.03.
As shown in Table~\ref{tab:sc}, we confirm through manual inspection that the given threshold indeed distinguishes overly simplified and semantically wrong bootcamps effectively.

\begin{table}[htb]
  \centering
  \caption{
  The proportions of different error types in LLM generated bootcamps with different self-consistent accuracy.
  We obtain self-consistent accuracy by using one bootcamp to generate problem instances and verify the responses of \llmname{Deepseek-R1-Distill-Qwen-32B}.
  }
  \vspace{0.2cm}
  \label{tab:sc}
  \resizebox{0.95\linewidth}{!}{
  \begin{tabular}{cccc}
    \toprule
    \textbf{Self-consistent Accuracy} & \textbf{Semantically Wrong} & \textbf{Overly Simplified} & \textbf{No Obvious Errors}\\
    \midrule
    $[0, 0.03)$ &  86.8\% & 0.0\% & 13.2\%\\
    $[0.03,0.85)$ &  18.8\% & 10.6\% & 70.6\%\\
    $[0.85, 1]$ &  0.0\% & 100.0\% & 0.0\%\\
    \bottomrule
  \end{tabular}
  }
\end{table}

By combining evolutionary refinement and self-consistent filtering, we successfully scale up the initial 100 manually implemented bootcamps to over 1000 automatically generated candidates. After applying heuristic strategies such as deduplication, difficulty calibration, and quality filtering, we retain 704 tasks for subsequent task scaling experiments, forming the core task set of \sandbox{}.

\subsection{Bootcamp Usage}
\label{sec:usage}

Leveraging the unified interface designed for \sandbox{}, our framework functions as an extensible library that can be seamlessly integrated into diverse language model applications.

\paragraph{BootcampEval} 
Leveraging the infinite reasoning problem generation capability of \texttt{case\_generator}, we constructed ~\eval{}, a large-scale benchmark for evaluating cross-domain reasoning capabilities in LLMs. This benchmark comprises 9,232 unique samples across 118 human-curated tasks distributed as described in Table \ref{tab:eval}. 
In all experiments, we ensure no data leakage by deduplicating the training data against \eval{}, guaranteeing clean evaluation and preventing train-test contamination. The complete benchmark suite is open-sourced.

\begin{wraptable}{r}{0.46\textwidth}  
  \centering
  \vspace{-25pt}
\caption{The distribution of tasks from \eval{}. All categories maintain uniform sampling (100 samples per task) except Cryptography, which has reduced sampling density due to strong domain correlation.}
\label{tab:eval}
  \resizebox{1\linewidth}{!}{
    \begin{tabular}{lcc}
    \toprule
    \textbf{Task Category} & \textbf{Task Count} & \textbf{Sample Count} \\ 
    \midrule
    Algorithm & 20 & 2,000 \\
    Character Puzzles & 6 & 600 \\
    Cryptography & 31 & 532 \\
    Graphical Puzzles & 31 & 3,100 \\
    Language Analysis & 3 & 300 \\
    Logical Reasoning & 16 & 1,600 \\
    Mathematical Modeling & 4 & 400 \\
    Natural Science & 7 & 700 \\ 
    \midrule
    \textbf{Total} & \textbf{118} & \textbf{9,232} \\ 
    \bottomrule
  \end{tabular}
  }
  \vspace{-30pt}  
\end{wraptable}

\paragraph{Data Synthesis and Rejection Finetuning}
Since \sandbox{} supports infinite query generation through parameterized templates and rule-based verification, it is particularly well-suited for data synthesis pipelines, especially rejection sampling-based methods~\cite{rejectionsampling}. Users can select multiple reasoning tasks from our codebase or extend it with custom tasks, then leverage \sandbox{} to distill strong language models for harvesting high-quality, diverse reasoning data for LLM training.

\paragraph{RLVR Training} 
\sandbox{} is natively compatible with RLVR (Reinforcement Learning with Verifiable Rewards) training due to its decoupled architecture built upon the unified interface. Our codebase already supports integration with RL training frameworks such as VeRL~\cite{verl} and XTuner~\cite{xtuner}. The only requirement for users is to import our \texttt{verify\_function} and transform the data format into specifications required by the target framework (e.g., Parquet format for VeRL), for which we provide ready-to-use conversion scripts. See Appendix~\ref{apd:integration} for a code example illustrating how to integrate \sandbox{} into RL training frameworks.

\section{Experiments: Boosting LLM Reasoning with Verifiable Task Scaling}

In our preliminary experiments, we compared model performance when trained on over 100 human-curated reasoning tasks versus training on single-domain tasks such as mathematics and coding. The results validate our hypothesis that multi-task training enhances LLMs' comprehensive reasoning capabilities and generalization performance. Building on these findings, we expanded our Bootcamp classes collection to 704 tasks through our automated agent synthesis workflow (Section \ref{sec:autogen}), which enables systematic investigation into how scaling the number of training tasks impacts both training efficiency and LLM reasoning proficiency.

\subsection{Setup Details}

To conduct task scaling experiments, we randomly selected task subsets containing 8, 32, 128, and 512 tasks from our task pool(704 tasks in total) and generated sufficient training data. While the random task selection may result in task categories overlapping with those in \eval{}, we ensure no data-level contamination by filtering all training instances to exclude any that share identities with \eval{}. Specifically, for the 8-task experiments, we conducted 4 independent trials with distinct task subsets and averaged the results to mitigate sampling variance. We selected \llmname{Qwen2.5-7B-Instruct}~\cite{qwen2_5} as the base model for its demonstrated general capabilities and relatively small parameter count, which enabled efficient execution of multiple experimental trials.

To eliminate potential biases from synthetic data sources in model training, we adopt the Zero-RL strategy to better validate the effectiveness of task scaling. A DAPO~\cite{dapo}-like RL algorithm was chosen for training. Our approach employs dynamic sampling during the rollout process, ensuring each training batch contains samples with verification scores between 0 and 1 to maintain a fixed effective training batch size. This mechanism thereby guarantees consistent data quantity across gradient updates, enabling direct comparison of training efficiency across steps. For DAPO configuration, we set the prompt batch size to 128 and sampled 8 responses per step with a temperature setting of 1. We did not use token-level loss in DAPO due to observed entropy instability during preliminary experiments. Each experiment was trained for a maximum of 500 steps.

We evaluated model performance on our \eval{} benchmark, which comprises 118 cross-domain reasoning tasks spanning 8 diverse domains. During both RL training and evaluation, we employed the training template illustrated in Appendix~\ref{apd:prompt}. This template requires the model to explicitly separate internal thought analysis (enclosed in "<think>" tags) from final answer generation, ensuring user-friendly reasoning outputs. To further enforce compliance with this format, we incorporated format penalties to validate proper "<think>" tag usage and length penalties to ensure sufficient reasoning depth before and after the thinking process.

\subsection{Main Results}

\label{sec:task_scaling}

\subsubsection{Training Efficiency and Reasoning Performance}

\begin{figure*}[htb]
    \centering
    \begin{subfigure}[b]{0.47\linewidth}
        \includegraphics[width=\linewidth, height=4.7cm, keepaspectratio]{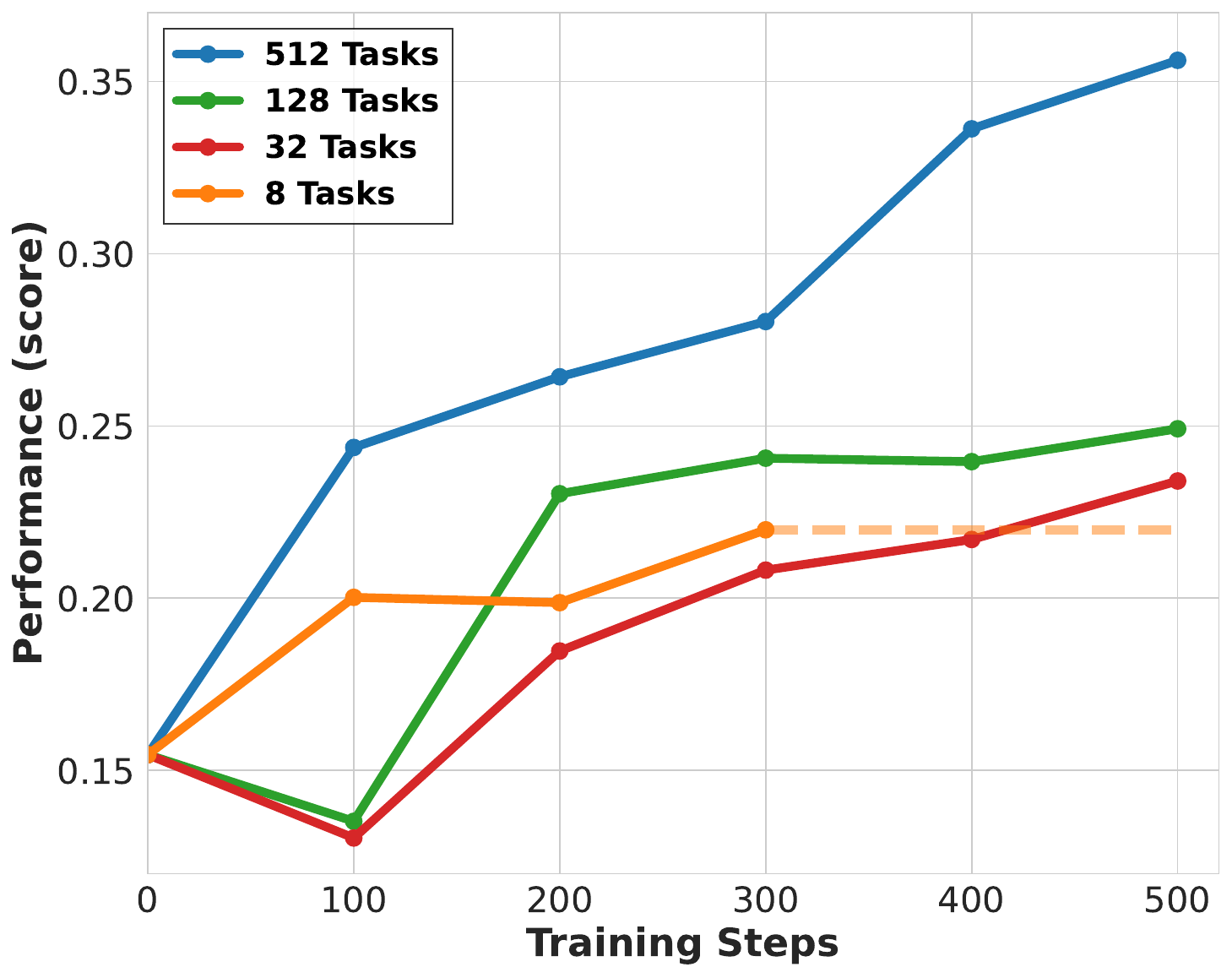}
        \caption{Evaluation performance on \eval{} with different numbers of training tasks.}
        \label{fig:taskscaling_overall_performance}
    \end{subfigure}
    \hfill
    \begin{subfigure}[b]{0.47\linewidth}
        \includegraphics[width=\linewidth, height=4.7cm, keepaspectratio]{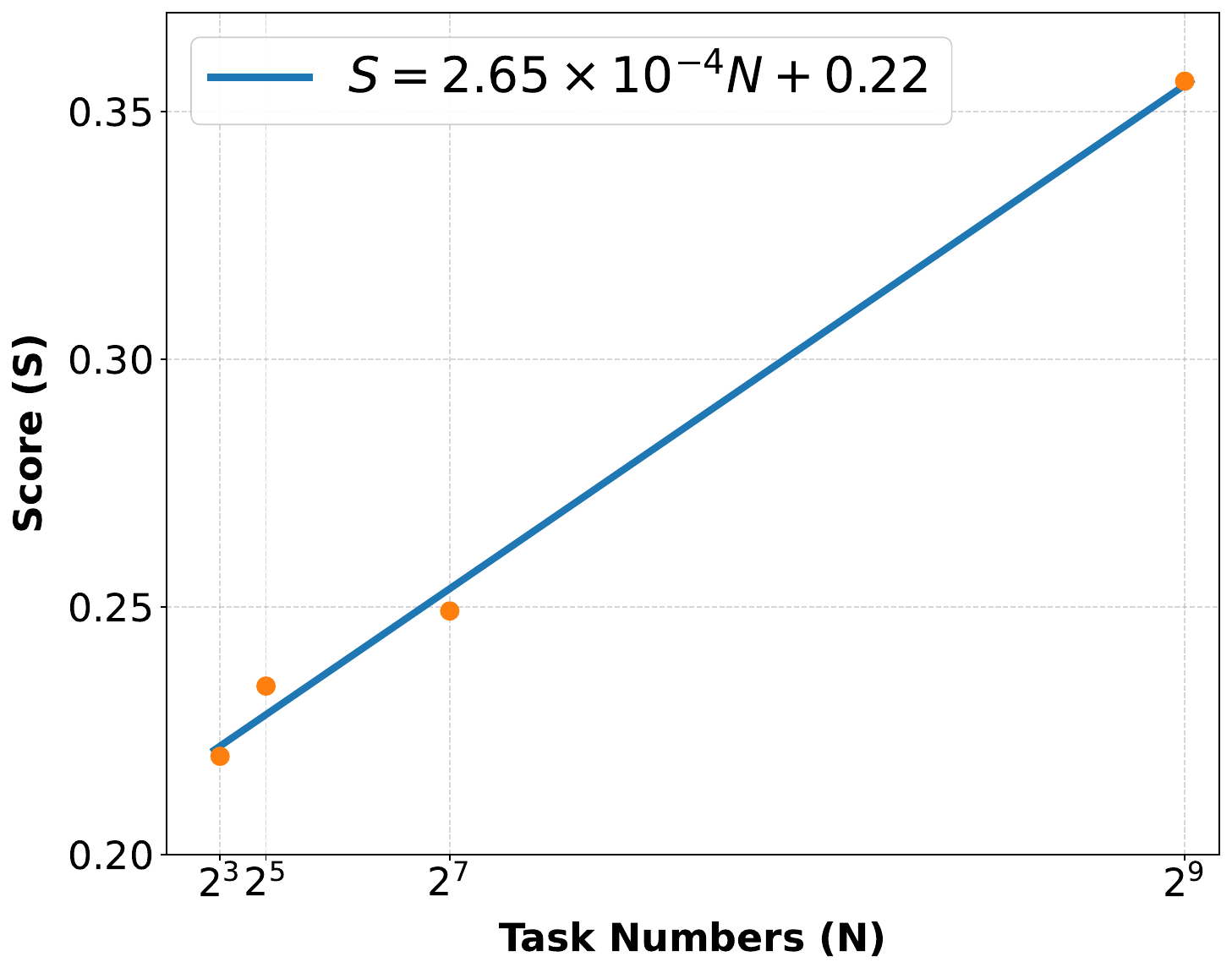}
        \caption{ Scaling trend between reasoning performance on \eval{} and the number of training tasks.}
        \label{fig:scaling_trend}
    \end{subfigure}
    

\caption{Scaling the number of training tasks improves reasoning performance in RL. 
(a) Performance on \eval{} increases with task count, demonstrating consistent gains in reasoning generalization. For the 8-task setting, the roll-out process collapses after 300 steps due to degenerate responses (Figure \ref{fig:efficiency}), resulting in no valid batches for evaluation. 
(b) A linear fit between task count and performance suggests a systematic improvement from task scaling.}
    \label{fig:mainresult}
    \vspace{-5pt}
\end{figure*}

\begin{wrapfigure}{r}{0.5\linewidth} 
    \vspace{-25pt}
  \includegraphics[width=\linewidth]{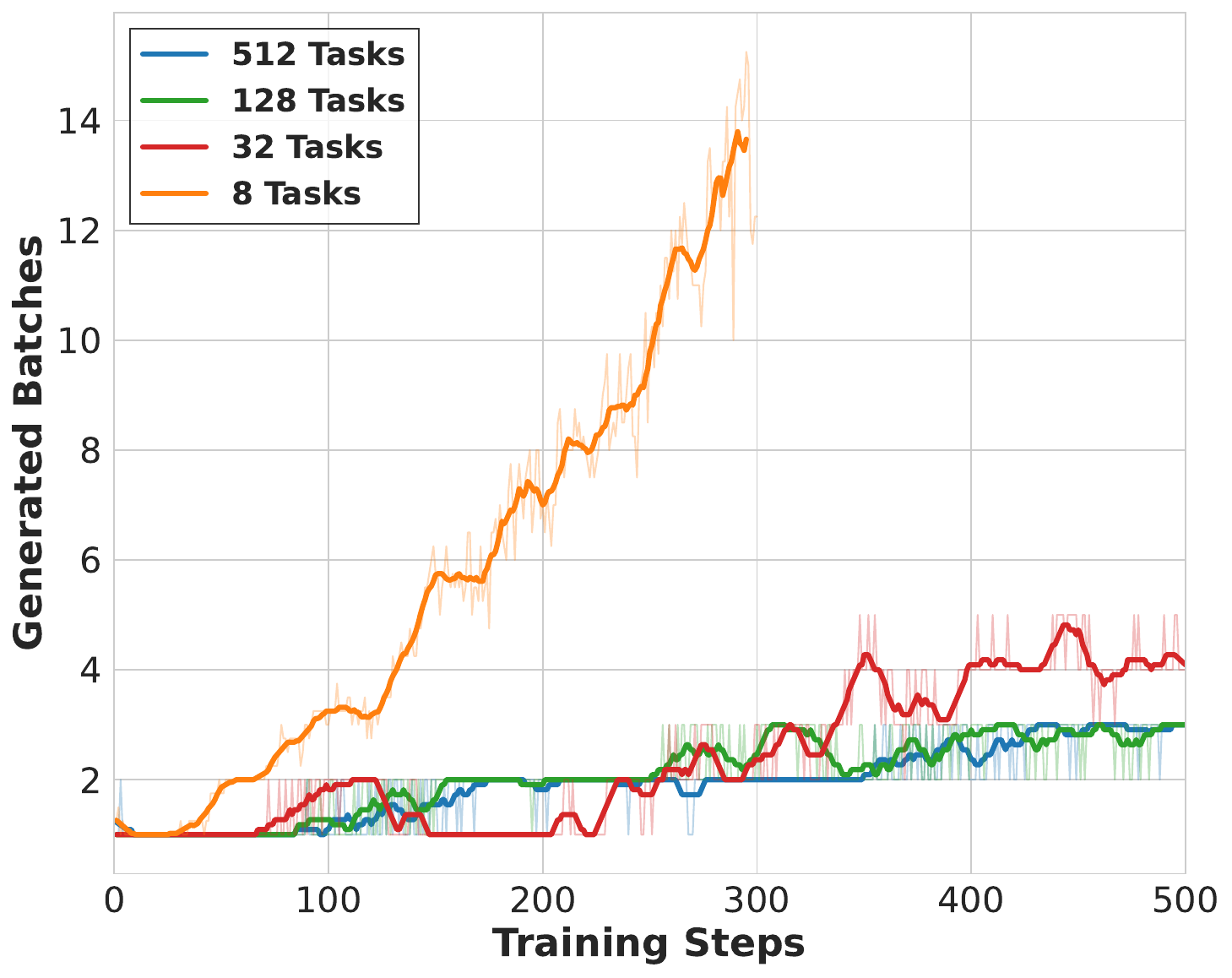}

\caption{With dynamic sampling in DAPO roll-out, the number of generated batches per training step increases over time. Training with more tasks avoids degenerate response patterns (e.g., all-correct or all-wrong), enabling stable value estimation, whereas few-task training leads to ineffective roll-out due to poor response diversity (entropy collapse).}
  \label{fig:efficiency}
    \vspace{-90pt}  
\end{wrapfigure}

We evaluated training efficiency and reasoning performance through task scaling experiments. Results in Figure~\ref{fig:mainresult} and~\ref{fig:efficiency} demonstrate that scaling the number of training tasks enhances both training efficiency and LLM reasoning capabilities. 

\paragraph{Training Efficiency} 
During the DAPO rollout process, the algorithm keeps sampling until the batch is fully filled with samples having verification scores strictly between 0 and 1. 
We adopted the oversampling strategy where the rollout batch size is three times(384) the prompt batch size(128). 
Consequently, we recorded the number of generated rollout batches per step (Figure~\ref{fig:efficiency}). 
Notably, the 8-task configuration exhibits a sharp increase in the number of generated batches as training progresses, indicating rapid performance plateauing – the model produces either entirely correct/incorrect responses, yielding insufficient preference data for optimization. 
This results in excessive invalid rollout computation, significantly compromising training efficiency. 
Therefore, RL training for the 8-task setting was terminated at step 300. 
In contrast, scaling up training tasks (32, 128, 512) shows minimal batch generation growth, suggesting RL progress maintains efficient sampling under diverse task distributions, thereby improving training efficiency.

\begin{figure*}[htb]
    \centering
    \includegraphics[width=\linewidth]{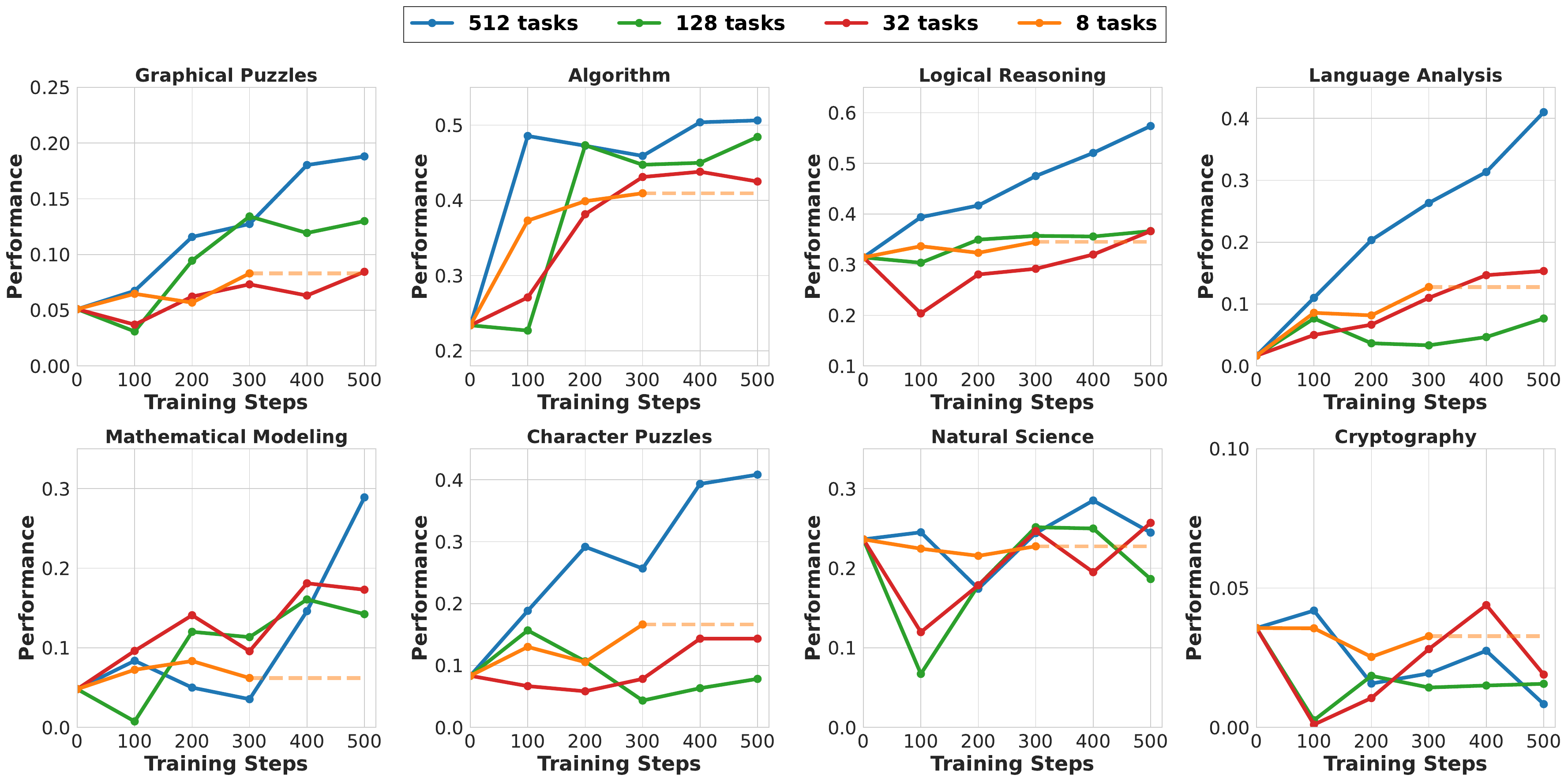}
    \caption{Detailed evaluation performance on different domains of \eval{}. We reported the average score of all our training checkpoints on 8 sub-domains of \eval{}.}

    \label{fig:performance_domain}
    \vspace{-5pt}
\end{figure*}

\paragraph{Reasoning Performance} Results in Figure~\ref{fig:taskscaling_overall_performance} demonstrate that scaling the number of training tasks significantly enhances the reasoning capabilities of large language models. With equal training steps, models trained on larger task sets (e.g., 512 tasks) outperform those trained with fewer tasks when evaluated on  \eval{}. Figure~\ref{fig:performance_domain} further illustrates the domain-specific performance across different configurations as training step increases. Notably, despite maintaining a fixed batch size (resulting in reduced per-task data exposure when scaling task numbers), models trained on diverse task distributions consistently achieve superior performance on most domains. Figure~\ref{fig:scaling_trend} reveals a strong positive correlation between reasoning performance and the quantity of training tasks. Specifically, the validation scores on \eval{} exhibit a near-linear improvement as the number of training tasks increases from 8 to 512. This observation empirically validates the effectiveness of task scaling in enhancing cross-domain reasoning capabilities.

\subsubsection{Multitask Training Enables Learning on Tasks Unsolvable in Isolation}

\begin{figure*}[htb]
    \centering
    \includegraphics[width=\linewidth]{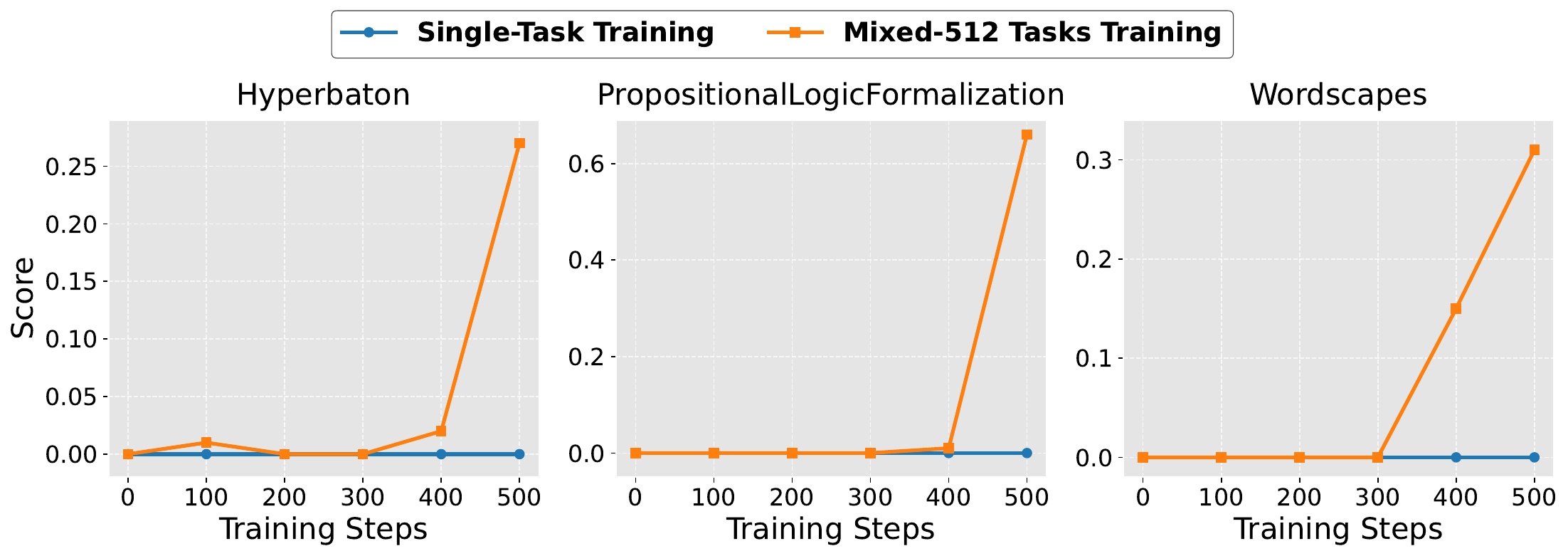}
    \caption{Emergent Moment. In our RL experiments, the 7B model fails to achieve performance gains when trained individually on these three tasks. However, significant improvements are observed after approximately 300 steps when the model is trained on a diverse mixture of 512 tasks. }

    \label{fig:emergent}
\end{figure*}

Notably, benefiting from intertask generalization capabilities, we observe that training on diverse task mixtures enables the emergence of latent reasoning abilities in tasks that fail to learn under isolated training conditions. As demonstrated in Figure~\ref{fig:emergent}, the model struggles to generate valid solutions for \texttt{Hyperbaton}, \texttt{PropositionalLogicFormalization}, and \texttt{Wordscapes} tasks when trained individually, resulting in stagnant learning curves. However, these capabilities exhibit a critical transition point after some steps of 512-tasks mixed RL training, where the model demonstrates significant performance improvements, which we refer as \textbf{"Emergent Moment"}. This phenomenon suggests that task scaling in RLVR through diverse task exposure fosters latent generalization capabilities, where cross-task knowledge transfer from abundant reasoning patterns enables progressive mastery of complex challenges. The observed performance demonstrate that scaling the number of training tasks inherently builds robust reasoning foundations, allowing LLMs to generalize across heterogeneous domains even without explicit curriculum design.

\subsection{Full Task Training with 32B Models}

Results in Section \ref{sec:task_scaling} demonstrate that scaling the number of verifiable tasks from \sandbox{} in RLVR training significantly improves both training efficiency and reasoning performance. This finding motivates us to conduct more experiments to fully utilize all 1000+ reasoning tasks in \sandbox{} for exploring the upper bound of model reasoning capability enhancement.

As detailed in Section \ref{sec:usage}, \sandbox{} supports both RLVR training and data synthesis. For SFT training, we first generate queries covering all 1000+ tasks from \sandbox{} and utilize \llmname{DeepSeek-R1} to synthesize a total of 55K long-CoT reasoning samples. Additionally, we collect 11K math data from open-source datasets~\cite{dapo,lightr1} as supplementary training sets. For RL training, we sample sufficient data from all \sandbox{} tasks and perform GRPO training with a prompt batch size of 256. We perform SFT for 3 epochs and RL for a maximum of 300 steps.

\begin{table}[htb]
    \tiny
 \renewcommand{\arraystretch}{1.03} 
  \centering
  \caption{
  Performance of SOTA Large Language Model and our trained LLMs on \eval{}.
  }
  \vspace{0.2cm}
  \label{tab:result_bootcamp_eval}
  \begin{tabular}{lc@{\hskip 5pt}c@{\hskip 5pt}c@{\hskip 5pt}c@{\hskip 5pt}c@{\hskip 5pt}c@{\hskip 5pt}c@{\hskip 5pt}c@{\hskip 5pt}c}
    \toprule
    \textbf{Model} & \begin{tabular}[c]{@{}c@{}}\textbf{Graph.}\\\textbf{Puzzles}\end{tabular} & \textbf{Algo} &  \textbf{Logic}   & \begin{tabular}[c]{@{}c@{}}\textbf{Crypto-}\\ \textbf{Graphy} \end{tabular} &  \textbf{Math} & \begin{tabular}[c]{@{}c@{}} \textbf{Character}\\ \textbf{Puzzles} \end{tabular} & \begin{tabular}[c]{@{}c@{}} \textbf{Natural} \\ \textbf{Science}  \end{tabular} & \begin{tabular}[c]{@{}c@{}} \textbf{Language}\\ \textbf{Analysis}  \end{tabular} & \textbf{Overall} \\
    \midrule
    \multicolumn{10}{c}{\textit{SOTA Open Source LLMs}} \\
    DeepSeek-V3-0324 &  33.0 & 57.4 & 50.4 & 36.2 & 50.7 & 75.2 & 46.8 & 37.0 & 46.5 \\
    DeepSeek-R1-0528 &  47.2 & 46.6 & 50.9 & \textbf{61.1} & 57.0 & 75.7 & 45.8 & 67.7 & 51.0 \\

    QwQ-32B &  45.2 & 53.8 & 56.7 & 37.6 & 60.4 & 72.5 & 42.3 & 44.7 & 51.4 \\
    Qwen3-32B &  45.9 & 53.7 & 53.7 & 45.8 & 56.4 & 81.5 & \underline{51.5} & 39.7 & 52.1 \\
    Qwen3-235B-A22B &  49.5 & 55.8 & 54.5 & 54.0 & 56.9 & 81.3 & 49.1 & 56.3 & 54.5 \\

    \midrule
    \multicolumn{10}{c}{\textit{Models Trained with InternBootcamp}} \\

    DS-R1-Distilled-Qwen-32B &  31.5 & 20.3 & 49.8 & 25.4 & 32.8 & 52.0 & 27.7 & 25.0 & 31.5 \\
    \rowcolor{lightgray}
    + Bootcamp-RL &  41.8 & \textbf{61.3} & \textbf{70.4} & 36.7 & \textbf{70.6} & 80.8 & 45.1 & 52.0 & 55.7 \\
    Qwen2.5-32B-Instruct &  13.4 & 30.4 & 43.3 & 12.4 & 10.9 & 24.2 & 29.9 & 10.0 & 24.4 \\
    \rowcolor{lightgray}
    + Bootcamp-RL &  25.7 & \underline{60.1} & \underline{67.4} & 18.5 & 36.8 & 59.5 & 48.4 & \underline{72.0} & 46.9 \\
    \rowcolor{lightgray}
    + Bootcamp-SFT &  \textbf{60.2} & 55.4 & 61.5 & \underline{56.0} & 60.3 & \textbf{87.7} & \textbf{56.6} & 69.3 & \textbf{61.1} \\
    \rowcolor{lightgray}
    + Bootcamp-SFT-RL &  \underline{52.4} & \textbf{61.3} & 64.1 & 47.5 & \underline{63.7} & \underline{87.5} & 44.4 & \textbf{77.3} & \underline{59.5} \\
    \bottomrule
  \end{tabular}

\end{table}

\begin{table}[htb]
 \tiny
 \renewcommand{\arraystretch}{1.05}
  \centering
  \caption{
  Performance of our trained 32B LLMs on OOD Reasoning Benchmarks.
  }
  \vspace{0.2cm}
  \label{tab:ood_eval}
  \begin{tabular}{l@{\hskip 8pt}c@{\hskip 5pt}c@{\hskip 5pt}c@{\hskip 5pt}c@{\hskip 5pt}c@{\hskip 5pt}c@{\hskip 5pt}c@{\hskip 5pt}c@{\hskip 5pt}c@{\hskip 5pt}c}
    \toprule

 & \multicolumn{2}{c}{\textbf{Logical Reasoning}} & \multicolumn{3}{c}{\textbf{Knowledge}}                                                                  & \multicolumn{2}{c}{\textbf{Math}} & \multicolumn{2}{c}{\textbf{Code}}                         &  \\ 

                      \cmidrule(lr){2-3} \cmidrule(lr){4-6} \cmidrule(lr){7-8} \cmidrule(lr){9-10} 
    \textbf{Model} & BBEH & \begin{tabular}[c]{@{}c@{}}KOR\\ Bench \end{tabular} &  \begin{tabular}[c]{@{}c@{}}GPQA\\ Diamond \end{tabular}   &  \begin{tabular}[c]{@{}c@{}}Super\\ GPQA \end{tabular} &  \begin{tabular}[c]{@{}c@{}}MMLU\\ Pro \end{tabular} &   \begin{tabular}[c]{@{}c@{}}AIME\\ 2025 \end{tabular}  & \begin{tabular}[c]{@{}c@{}} LMB\\ Hard \end{tabular} & \begin{tabular}[c]{@{}c@{}} Human\\ Eval \end{tabular} & \begin{tabular}[c]{@{}c@{}} LCB\\ v6  \end{tabular} & \textbf{Avg.} \\
    \midrule
    DS-R1-Distilled-Qwen-32B &  27.8 & 70.7 & 41.6 & 45.9 & 73.0  & 52.5 & 36.8 & 90.2 & 33.7 & 52.5 \\
    \rowcolor{lightgray}
    + Bootcamp-RL &  27.0 & 73.5 & \underline{51.6} & \underline{48.7} & 70.6  & \underline{56.8} & \underline{43.7} & \underline{95.6} & \underline{44.6} & \underline{56.9} \\
    Qwen2.5-32B-Instruct &  19.4 & 60.7 & 44.7 & 39.3 & 69.1  & 10 & 22.0 & 87.2 & 28.0 & 42.3\\
    \rowcolor{lightgray}
    + Bootcamp-RL &  20.3 & 60.1 & 44.7 & 38.0 & 68.3  & 14.7 & 21.8 & 89.6 & 29.7 & 43.0\\
    \rowcolor{lightgray}
    + Bootcamp-SFT &  \textbf{36.7} & \textbf{79.3} & 26.2 & 48.5 & \underline{76.1}  & 41.2 & 33.2& 93.9 & 44.0 & 53.2\\
    \rowcolor{lightgray}
    + Bootcamp-SFT-RL &  \underline{35.9} & \underline{78.9} & \textbf{60.7} & \textbf{51.0} & \textbf{77.1} & \textbf{60.5} & \textbf{46.2} & \textbf{98.8} & \textbf{47.4} & \textbf{61.8}\\
    \bottomrule
  \end{tabular}

\end{table}

To comprehensively verify and compare the effects of training with the full set of Bootcamp tasks as an ablation study, we chose two baselines and conducted four experimental setups. For \llmname{DeepSeek-R1-Distill-Qwen-32B} (a model already fine-tuned with long CoT reasoning data), we performed RL training directly using all tasks from \sandbox{}. For \llmname{Qwen2.5-32B-Instruct} (a strong baseline for training reasoning models), we employed three separate approaches: (a) RL training only, (b) SFT training only, and (c) a combination of SFT followed by RL training.

We evaluated our trained models on our in-domain benchmark \eval{}, as well as several challenging out-of-domain reasoning benchmarks covering Logical Reasoning~\cite{BBEH,korbench}, Knowledge Reasoning~\cite{GPQA,supergpqa}, Mathematical Reasoning~\cite{aime,livemathbench}, and Code Generation~\cite{humaneval,livecodebench}. Results are summarized in Table~\ref{tab:result_bootcamp_eval} (in-domain) and Table~\ref{tab:ood_eval} (out-of-domain).

As shown in Table~\ref{tab:result_bootcamp_eval}, models trained on data generated from \sandbox{} outperform open-source state-of-the-art LLMs on \eval{}, validating the effectiveness of our methods. Furthermore, Table~\ref{tab:ood_eval} demonstrates substantial performance gains on several out-of-domain benchmarks compared to baselines. This indicates that \sandbox{} enhances comprehensive reasoning capabilities and enables effective generalization to diverse reasoning domains.

Experimental results in Table~\ref{tab:result_bootcamp_eval} and~\ref{tab:ood_eval} further demonstrate the effectiveness of \sandbox{} across both RFT and RLVR paradigms. Notably, sequential application of SFT followed by RL achieves the maximum performance gains. For \llmname{DeepSeek-R1-Distilled-Qwen-32B} (already finetuned with long-CoT reasoning data), directly applying RL training with \sandbox{} tasks yields significant improvements on both in-domain and OOD benchmarks. For another baseline \llmname{Qwen2.5-32B-Instruct} without prior long-CoT training, standalone RL and SFT both enhance reasoning capabilities, though SFT exhibits substantially stronger gains. Critically, applying RL after SFT on \sandbox{} data produces the best-performing model, which not only achieves superior scores in mathematical/code domains but also demonstrates unexpected improvements in knowledge-intensive reasoning benchmarks (GPQA Diamond, SuperGPQA, MMLU Pro). Since reinforcement learning inherently cannot inject new knowledge, we posit that scaling tasks in RL training better activates the model's existing knowledge application and generalization capabilities.

\section{Conclusion}

In this work, we present \sandbox{}, the first large-scale extensible library of environments for training large reasoning models. It integrates 1000+ general reasoning tasks across 8 domains with difficulty-controllable generators and rule-based verifiers, supporting both RLVR training and data synthesis. We also introduce the \eval{} benchmark, covering 118 cross-domain reasoning tasks to holistically assess the general reasoning capabilities of LLMs. Our experiments demonstrate the effectiveness of verifiable task scaling in boosting training efficiency and reasoning performance. With the combination of SFT and RLVR with data synthesized from \sandbox{}, our trained 32B model outperforms SOTA open-source LLMs on \eval{} and delivers significant performance gains on OOD reasoning benchmarks. We open-source \sandbox{} to serve the community, fostering deeper exploration of how task scaling enhances LLM reasoning generalization.

\section{Acknowledgement}
This work was supported by Shanghai Artificial Intelligence Laboratory.

\newpage
\appendix

\section{Task Sources and Curation}
\label{apd:task_sources}
We notice that there exists a wide range of reasoning tasks in the real world that are verifiable.
We believe that these tasks cover diverse reasoning behaviors and would be helpful to build versatile reasoners.
To collect these tasks, we first identify several representative task categories and then extensively collect task descriptions from web pages or datasets where these task categories are densely concentrated.
We detail some main task sources and their features as follows.

\textbf{Puzzles} are a class of synthetic reasoning tasks defined by a series of given rules.
Representative puzzles include \textit{Sudoku}, \textit{Minesweeper}, \textit{Star Battle}, \textit{Cryptarithms}, \textit{Arrow Maze}, \textit{Binairo}, etc.
These tasks offer flexible difficulty tuning by adjusting problem sizes, and their solutions encompass diverse reasoning patterns, namely deductive, inductive, and abductive reasoning~\citep{giadikiaroglou2024puzzle}. 
Consequently, they have long been used to test or foster children’s ingenuity~\citep{zeinalipour2023italian} and, more recently, to evaluate the reasoning abilities of large language models~\citep{yao2023tree,gandhistream,lin2025zebralogic,mittal2024puzzlebench,giadikiaroglou2024puzzle,chia2024puzzlevqa,ghosal2024algopuzzlevqa}. To compile our dataset of puzzle tasks, we drew on a series of puzzle websites\footnote{\url{https://www.puzzle-nonograms.com}} developed by PuzzleTeam\footnote{\url{https://www.youtube.com/channel/UCiU0kY99jtY-iQpzulYk8JQ}}, and scraped task descriptions from these sites. 
Given the brevity of the original descriptions, we further enriched them by incorporating knowledge from LLMs (e.g., \llmname{DeepSeek-V3}~\cite{dsv3}) to add details of the task rules.

\textbf{Reasoning Benchmarks} covers a number of widely adopted benchmarks to evaluate the reasoning abilities of large language models.
Currently, we include tasks from ARC-AGI\footnote{\url{https://arcprize.org/arc-agi}}, KOR-Bench~\citep{korbench}, and BBEH~\citep{BBEH}. 
We consider these tasks in our collection to help study reasoning capabilities that the community is interested in.
Specifically, KOR-Bench includes five types of reasoning tasks: logic, operation, cipher, puzzle, and counterfactual reasoning, where we neglect counterfactual reasoning for its dependence on specific world-view knowledge and build bootcamps for the remaining four types of tasks. 
BBEH contains 23 reasoning tasks obtained by complicating tasks from BBH, and we build bootcamps for tasks that do not depend on external knowledge.

\textbf{Algorithm Problems} encompass a variety of practical reasoning patterns defined by concrete algorithms, serving as ideal materials for studying generalization capabilities across tasks with similar reasoning patterns. 
Despite differences in the descriptions of different problems, their solutions can all be reduced to certain algorithms. 
Therefore, with abundant algorithm problems, we can study the generalization of reasoning models under well-defined settings by exploring how the models perform after learning on problems that share the same algorithms.
To collect these problems, we make use of CodeContest~\citep{doi:10.1126/science.abq1158}, a competitive programming dataset and it is straightforward to include more by considering similar datasets in the future.


\textbf{Scientific Tasks} represent a spectrum of reasoning-intensive endeavors deeply intertwined with scientific research activities, which are regarded as one of the most valuable domains where AI will revolutionize~\citep{wang2023scientific,lu2024ai,gottweis2025towards}.
We consider that improving reasoning models on these tasks facilitatesthe  achievement of this vision.

After collecting these tasks, we reorganize them according to their subjects.
The complete composition of all supported tasks in \sandbox{} is shown in Figure~\ref{fig:bootcamp_example}. 
The original description of each task is provided as comments in its corresponding bootcamp code within the open-source repository.





\section{Framework Design and Implementation}
\label{apd:framework}



\begin{figure*}[tb]
    \centering
    \includegraphics[width=\linewidth]{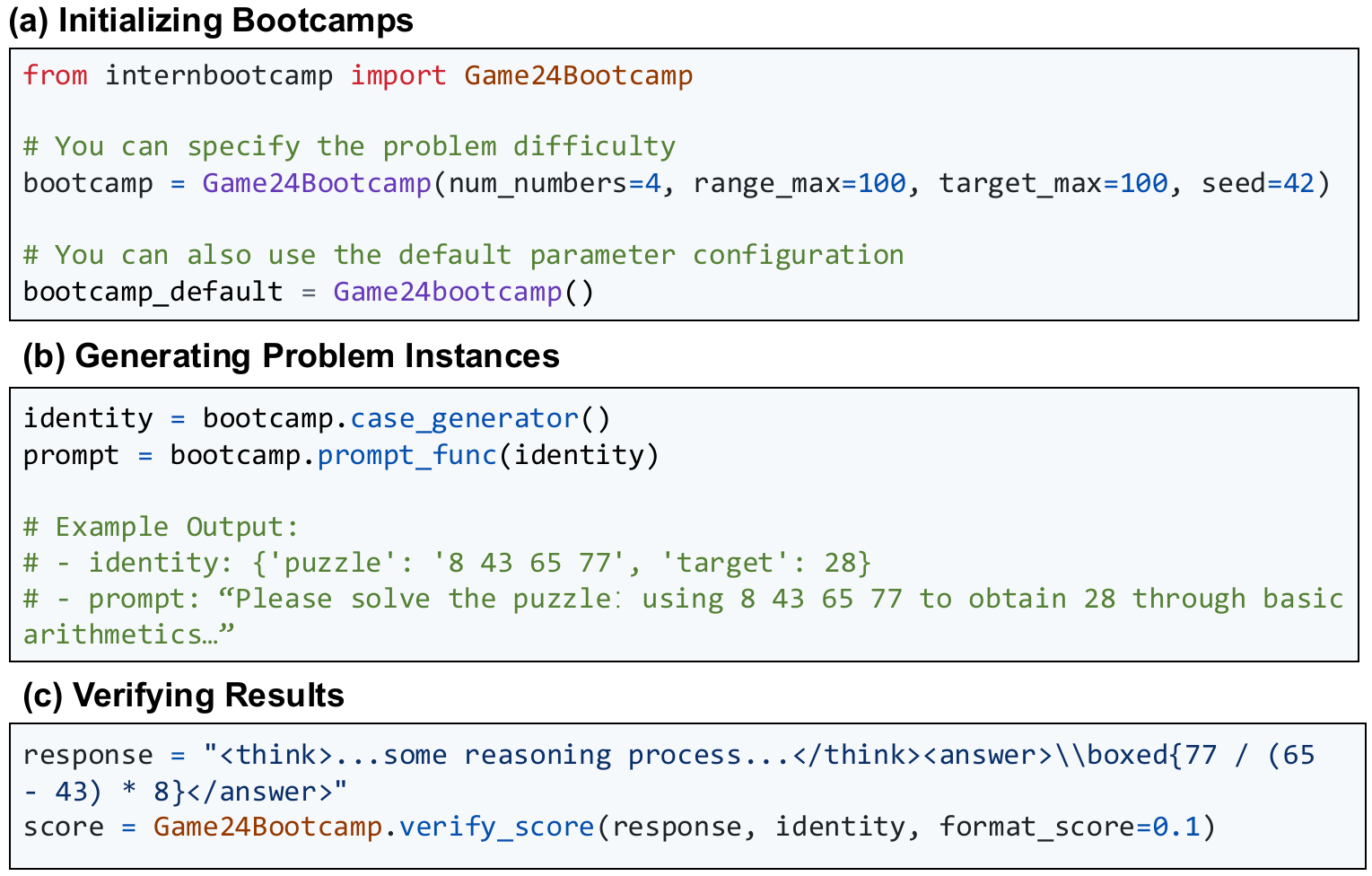}
    \caption{Example usage of the interfaces of \sandbox{} for \texttt{Game24}. (a) We first initialize a \texttt{Game24Bootcamp} by either specifying the configurations for the bootcamps, which control the difficulties of the produced problem instances, or using a default configuration(from a configuration file). (b) We use \texttt{case\_generator} interface to randomly generate the identity describing a problem instance, which is fed into the \texttt{prompt\_func} interface to transform into a natural language problem. (c) Given a language model response, we feed it with the corresponding identity to the \texttt{verify\_score} interface to verify the correctness of this response.}
    \label{fig:usage_example}
\end{figure*}

\sandbox{} adopts a modular architecture based on the \texttt{BaseBootcamp} abstract base class. Each task is implemented as a concrete Bootcamp class that inherits from \texttt{BaseBootcamp} and implements three core methods: \texttt{case\_generator} for generating problem instances, \texttt{prompt\_function} for formatting instances into prompts, and \texttt{verify\_function} for verifying solution correctness. An example of their usage is shown in Figure~\ref{fig:usage_example}.

\paragraph{(i) \texttt{case\_generator}} Each verifiable reasoning task in \sandbox{} implements a \texttt{case\_generator} method that instantiates concrete problems from abstract task definitions. Specifically, the generator reads parameters from the configuration file and randomly generates all necessary information to construct one query instance and its corresponding verification protocol, termed an "identity". For instance, in numeral tasks like \textit{Game24}, the generator reads the number of numerical operands and value ranges specified by the user from the configuration, then randomly generates a set of numerical values and a target value. For board-like tasks such as \textit{ArrowMaze}, it parses the board dimensions and required step counts from the configuration to synthesize a final maze layout. The identity encapsulates all information required to describe a task instance and verify its correctness.\footnote{Note that not all problems have unique solutions; thus, we use "identity" to represent the complete problem specification for verification purposes.}

\paragraph{(ii) \texttt{prompt\_function}:} After \texttt{case\_generator} generates the identity of one task instance, the \texttt{prompt\_function} constructs a query by formatting the identity into structured prompts. During implementation, we incorporated multi-lingual templates to generate language-specific prompts for the same identity, while ensuring that standardized answer formats are specified in the prompt instructions. This design enables a consistent and unambiguous evaluation of model responses by subsequent verification modules.

\paragraph{(iii) \texttt{verify\_function}:} The \texttt{verify\_function} method receives the LLM's raw output and the task identity. It first extracts the candidate answer from the model's response, then evaluates its correctness against the ground truth encoded in the identity. Finally, it computes a verification score ranging from 0 (completely incorrect) to 1 (fully correct), quantifying the alignment between the model's reasoning process and the expected solution. Moreover, we incorporate format penalties and length penalties in the verification function to identify malformed responses during RL training. Specifically, these mechanisms are designed to regularize model output patterns by penalizing deviations from expected structural constraints. Empirical results demonstrate that these improvements effectively enforce structured response generation while maintaining reasoning accuracy.

Additionally, each bootcamp class accepts an user-defined configuration file during initialization, which encodes hyperparameters for difficulty control and initialization. This design enables systematic exploration of task complexity.

\section{Integration with RL Training Frameworks}
\label{apd:integration}

To facilitate adoption in reinforcement learning pipelines, \sandbox{} provides modular support for RL training frameworks. The integration requires only a lightweight modification: plugging in the \texttt{verify\_score} function from the corresponding bootcamp as the reward computation module. As shown in Figure~\ref{fig:verl}, this can be achieved by routing task-specific verification through a unified scoring interface, enabling seamless compatibility with existing RL frameworks.

\begin{figure*}[ht]
\begin{lstlisting}
def default_compute_score(data_source, solution_str, ground_truth):
    ...
    # Modification: Add Bootcamp-specific verification
    elif data_source.startswith("bootcamp/"):    
        import importlib
        import json
        bootcamp_name = data_source.split("/")[1]
        class_name = bootcamp_name[0].upper() + bootcamp_name[1:] + "bootcamp"
        module = importlib.import_module("internbootcamp")
        ground_truth = json.loads(ground_truth)
        score = getattr(module,class_name).verify_score(solution_str,ground_truth)
        return score
    ...
    
\end{lstlisting}
\caption{An example of integrating \sandbox{} with RL frameworks like VeRL.}
\label{fig:verl}
\end{figure*}

\section{Training Prompt Template}
\label{apd:prompt}

\begin{figure}[htbp]  
    \centering 
\begin{tcolorbox}[
    enhanced,
    colframe=deepblue,
    colback=lightblue!20,
    boxrule=1pt,
    arc=2mm,
    title={Training Prompt Template},
    fonttitle=\bfseries,
    top=4pt,
    bottom=4pt,
    left=4pt,
    right=4pt
]
You are a helpful assistant, skilled at solving various complex reasoning problems. When faced with any user questions, please first conduct a detailed thinking process, similar to drafting, where you can freely analyze problem-solving strategies and verify the correctness of your thought process. Please put your thinking process within <think> and </think> tags. After completing the thinking process, provide the user with a detailed response. Please note that the response accessible to the user will start after </think>, so ensure that detailed chain-of-thought solution steps are provided after the </think> tag.
\end{tcolorbox}
    \caption{The prompt template used for training our models.}
    \label{fig:prompt}  
\end{figure}

%
%
%
\bibliographystyle{splncs04}
\bibliography{citations}
%




\end{document}